\documentclass[sigconf]{acmart}

\AtBeginDocument{%
  }

\copyrightyear{2026}
\acmYear{2026}
\setcopyright{cc}
\setcctype{by}
\acmConference[ICCAD '26]{IEEE/ACM International Conference on Computer-Aided Design}{November 08--12, 2026}{San Jose, CA, USA}
\acmBooktitle{IEEE/ACM International Conference on Computer-Aided Design (ICCAD '26), November 08--12, 2026, San Jose, CA, USA}
\acmDOI{10.1145/3831252.3834191}
\acmISBN{979-8-4007-2873-0/2026/11}
\usepackage{booktabs}
\usepackage{amsmath}
\usepackage{enumitem}
\usepackage{latexsym}
\usepackage{algorithm}
\usepackage{algpseudocode}
\usepackage{tabularx}
\usepackage{makecell}
\usepackage{graphicx}
\usepackage{subcaption}
\usepackage{multirow}
\usepackage{array}
\usepackage{microtype}
\usepackage{tcolorbox}
\usepackage{tikz}
\usetikzlibrary{arrows.meta,backgrounds}

\usepackage{xcolor}

\begin{document}

%%
%% The "title" command has an optional parameter,
%% allowing the author to define a "short title" to be used in page headers.
\title{HLS-Seek: QoR-Aware Code Generation for High-Level Synthesis via Proxy Comparative Reward Reinforcement Learning}

%%
%% The "author" command and its associated commands are used to define
%% the authors and their affiliations.
\author{Qingyun Zou}
\email{qingyunzou@u.nus.edu}
\affiliation{%
  \institution{National University of Singapore}
  \city{Singapore}
  \country{Singapore}
}

\author{Feng Yu}
\email{yuf@u.nus.edu}
\affiliation{%
  \institution{National University of Singapore}
  \city{Singapore}
  \country{Singapore}
}

\author{Hongshi Tan}
\email{hongshi@u.nus.edu}
\affiliation{%
  \institution{National University of Singapore}
  \city{Singapore}
  \country{Singapore}
}

\author{Yao Chen}
\email{yaochen\_nk@hotmail.com}
\affiliation{%
  \institution{Huazhong University of Science and Technology}
  \city{Wuhan, Hubei}
  \country{China}
}

\author{Bingsheng He}
\email{dcsheb@nus.edu.sg}
\affiliation{%
  \institution{National University of Singapore}
  \city{Singapore}
  \country{Singapore}
}

\author{Weng-fai Wong}
\email{wongwf@nus.edu.sg}
\affiliation{%
  \institution{National University of Singapore}
  \city{Singapore}
  \country{Singapore}
}

\renewcommand{\shortauthors}{Zou et al.}

%%
%% The abstract is a short summary of the work to be presented in the article.
\begin{abstract}
High-Level Synthesis (HLS) compiles algorithmic C/C++ descriptions into hardware, with Quality of Results (QoR)---latency and resource utilization---critically governed by pragma configurations and code structure. Existing natural-language-to-HLS (NL-to-HLS) training approaches prioritize functional correctness while largely ignoring QoR. We observe that reinforcement learning (RL) for HLS does not require absolute synthesis results---only relative comparisons between candidates. Based on this insight, we propose \textbf{HLS-Seek}, a QoR-aware NL-to-HLS framework that avoids full synthesis-in-the-loop RL via a comparative proxy reward model achieving 99.53\% Pareto-dominance accuracy. To prevent reward hacking, we introduce \textit{uncertainty-aware Monte Carlo (MC) dropout switching} that selectively invokes real Vitis HLS synthesis for low-confidence candidates and online updates the proxy, creating a self-improving reward system. HLS-Seek achieves 84.7\% syntax correctness pass@1 and 81.4\% functional correctness pass@5 on HLS-Eval~\cite{abikaram2025hlseval} with only 7B parameters, surpassing GPT-5.1 on functional pass@5, while achieving 8.5$\times$ faster training than real-reward RL. On QoR evaluation, HLS-Seek achieves the lowest latency on 19/30 kernels and Pareto-dominates HLS-specific baselines on 9 kernels.
\end{abstract}

%%
%% Keywords. The author(s) should pick words that accurately describe
%% the work being presented. Separate the keywords with commas.
\keywords{HLS code generation, reinforcement learning, quality of results, large language models}

\begin{CCSXML}
<ccs2012>
<concept>
<concept_id>10010583.10010682.10010684</concept_id>
<concept_desc>Hardware~High-level and register-transfer level synthesis</concept_desc>
<concept_significance>500</concept_significance>
</concept>
</ccs2012>
\end{CCSXML}

\ccsdesc[500]{Hardware~High-level and register-transfer level synthesis}

%%
%% This command processes the author and affiliation and title
%% information and builds the first part of the formatted document.
\maketitle

% \begin{figure}[b]
%   \centering
%   \includegraphics[width=\linewidth]{figs/reward_compared.pdf}
%   \caption{Comparison of RL training paradigms for HLS. Preference learning fails because minimal pragma token differences (e.g., \texttt{unroll factor=1} vs. \texttt{4}) can cause large QoR gaps. Real-reward RL is accurate but slow due to synthesis latency. Our hybrid reward combines lightweight signals (reasoning format, compilation, functional correctness) with a fast predicted QoR reward, achieving both larger reward discrimination and real-time feedback.}
%   \label{fig:reward_model}
% \end{figure}

\section{Introduction}

HLS compiles algorithmic C/C++ descriptions into Register Transfer Level (RTL) hardware, enabling rapid prototyping on field-programmable gate array (FPGA) and application-specific integrated circuit (ASIC) platforms~\cite{coussy2010high,lahti2018we,cong2022fpga}. It supports diverse domains, including machine-learning accelerators~\cite{duarte2018hls4ml,zeng2024flightllm,chen2019cloud}, graph processing~\cite{chen2021thundergp,dai2017foregraph,tan2023lightrw}, and data-intensive applications~\cite{mueller2009data,jiang2023data,alonso2020tackling}.

\begin{figure}[H]
  \centering
  \includegraphics[width=0.92\linewidth]{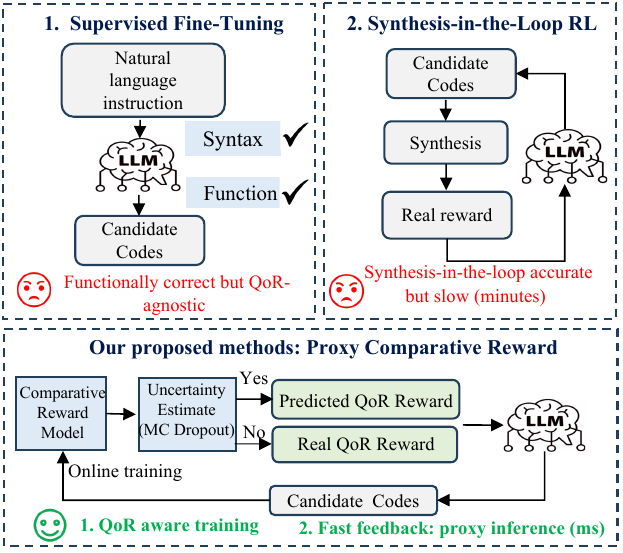}
  \Description{Comparison of three RL training paradigms for HLS code generation.}
\caption{Three training methods for LLM-based HLS code generation.}
%\td{remove DPO, replace with figure shows existing HLS is not QoR aware (one-short?) done }}
  \label{fig:reward_model}
\end{figure}

Recently, there has been growing interest in leveraging LLMs to automate HLS code generation from high-level specifications~\cite{gai2025exploring,khan2025sagehls,sheikholeslam2024synthai,swaroopa2024evaluating,peng2025forgehls}. For example, Gai et al.~\cite{gai2025exploring} fine-tune CodeLlama on HLS benchmarks for functional correctness, and SAGE-HLS~\cite{khan2025sagehls} incorporates abstract syntax tree (AST)-guided representations to improve code structure awareness, demonstrating that LLMs can generate synthesizable hardware code with increasing reliability. These efforts highlight the potential of LLMs to significantly improve developer productivity and lower the barrier to hardware design.

However, these approaches primarily aim to generate runnable kernels, while the QoR, which is critical for real-world accelerator deployment~\cite{cong2022fpga}, is largely overlooked. QoR reflects low-level hardware performance and efficiency, and is typically measured by execution latency and resource utilization, including lookup tables (LUTs), digital signal processing (DSP) blocks, block RAMs (BRAMs), and flip-flops (FFs). It is determined by the microarchitectural structure of the kernel, the pragma configurations and code optimizations applied in the HLS compilation flow, as well as hardware resource instantiation behavior and downstream physical implementation optimizations~\cite{cong2022fpga}. As a result, QoR cannot be accurately inferred from software code analysis alone and instead requires evaluation through the synthesis and implementation.

As shown in Figure~\ref{fig:reward_model}, existing training paradigms face limitations for QoR-aware HLS generation. Supervised Fine-Tuning (SFT)-based approaches~\cite{gai2025exploring,khan2025sagehls} produce functionally correct code but are QoR-agnostic---all compilable designs receive equal training signal regardless of hardware performance. In the RTL domain, ChipSeek~\cite{chen2025chipseek} demonstrates that real synthesis feedback can serve as an effective RL reward. However, extending this synthesis-in-the-loop paradigm to HLS is prohibitively expensive: each HLS synthesis run takes minutes to hours (e.g., ThunderGP~\cite{chen2021thundergp}), and RL training requires over 12,500 policy-update steps, each scoring a group of four candidates, making it infeasible at scale.

To overcome this bottleneck, we observe a key insight: \textit{RL training does not require absolute QoR values from synthesis---only relative comparisons between candidates.} Knowing which of two designs is better suffices for policy improvement via group-relative advantage estimation, without needing exact latency or resource numbers. This motivates us to train a lightweight \textit{comparative reward model} that predicts, given two candidate HLS designs, which achieves a better resource-latency trade-off---achieving 99.53\% dominance accuracy and 99.42\% latency accuracy without invoking synthesis. To safeguard against reward hacking as the RL policy explores out-of-distribution pragma configurations, the reward model estimates its own confidence via Monte Carlo (MC) dropout and selectively falls back to real Vitis HLS synthesis for high-uncertainty candidates. Crucially, these synthesis results are fed back to \textit{online update} the reward model, continuously expanding its coverage. This self-improving mechanism reduces the synthesis trigger rate over the course of training, keeping overall cost close to a proxy-only baseline while maintaining reward fidelity.

Based on this reward design, we propose \textbf{HLS-Seek}, a QoR-aware NL-to-HLS code generation framework trained in three stages. To address training data scarcity, we build upon the open-source dataset and apply a novel \textit{Pareto-Proximal Quality-Diversity Sampling} strategy to curate QoR-competitive HLS implementations from its design space exploration (DSE) results. We further leverage commercial reasoning models to generate chain-of-thought \textit{reasoning traces}, step-by-step rationales for code optimization. Training then proceeds as: (\textit{i}) diversity-oriented SFT on a broad HLS corpus with QoR descriptions, exposing the model to diverse pragma configurations; (\textit{ii}) cold-start SFT on Pareto-proximal data with reasoning traces, instilling hardware reasoning capability before RL; and (\textit{iii}) Group Relative Policy Optimization (GRPO)~\cite{shao2024deepseekmath} with a four-component reward---reasoning format, compilation, functional correctness, and predicted QoR from the comparative reward model.
Our main contributions are:
\begin{itemize}[leftmargin=1em, labelsep=0.5em]
\item We identify that RL for HLS code generation requires only relative QoR signals, not absolute synthesis results. Based on this insight, we propose a comparative proxy reward model with uncertainty-aware MC dropout switching that provides fast, accurate, and self-improving reward feedback during GRPO training.

\item We present \textbf{HLS-Seek}\footnote{\url{https://github.com/zanehpn/HLS_code_generations}}, a complete QoR-aware NL-to-HLS framework comprising: a three-stage training pipeline (diversity SFT $\to$ cold-start SFT with reasoning traces $\to$ proxy reward GRPO), a Siamese comparative reward model that provides fast pairwise QoR ranking, and an uncertainty-aware MC dropout switching mechanism that online updates the proxy with real synthesis.

\item HLS-Seek achieves 84.7\% syntax correctness pass@1 and 81.4\% functional correctness pass@5 on HLS-Eval with only 7B parameters, surpassing GPT-5.1 on functional pass@5 (81.4\% vs.\ 79.1\%) and leading all frontier and code-specialist baselines at pass@5 and pass@10 on the ForgeHLS test set. On QoR evaluation, HLS-Seek achieves the lowest latency on 19/30 kernels, Pareto-dominates SAGE-HLS on 5 kernels and Gai et al.\ on 4 kernels with only 1 reverse, and demonstrates algorithmic code restructuring beyond pragma insertion.
\end{itemize}

\begin{figure*}[t]
  \centering
  \includegraphics[width=0.80\textwidth]{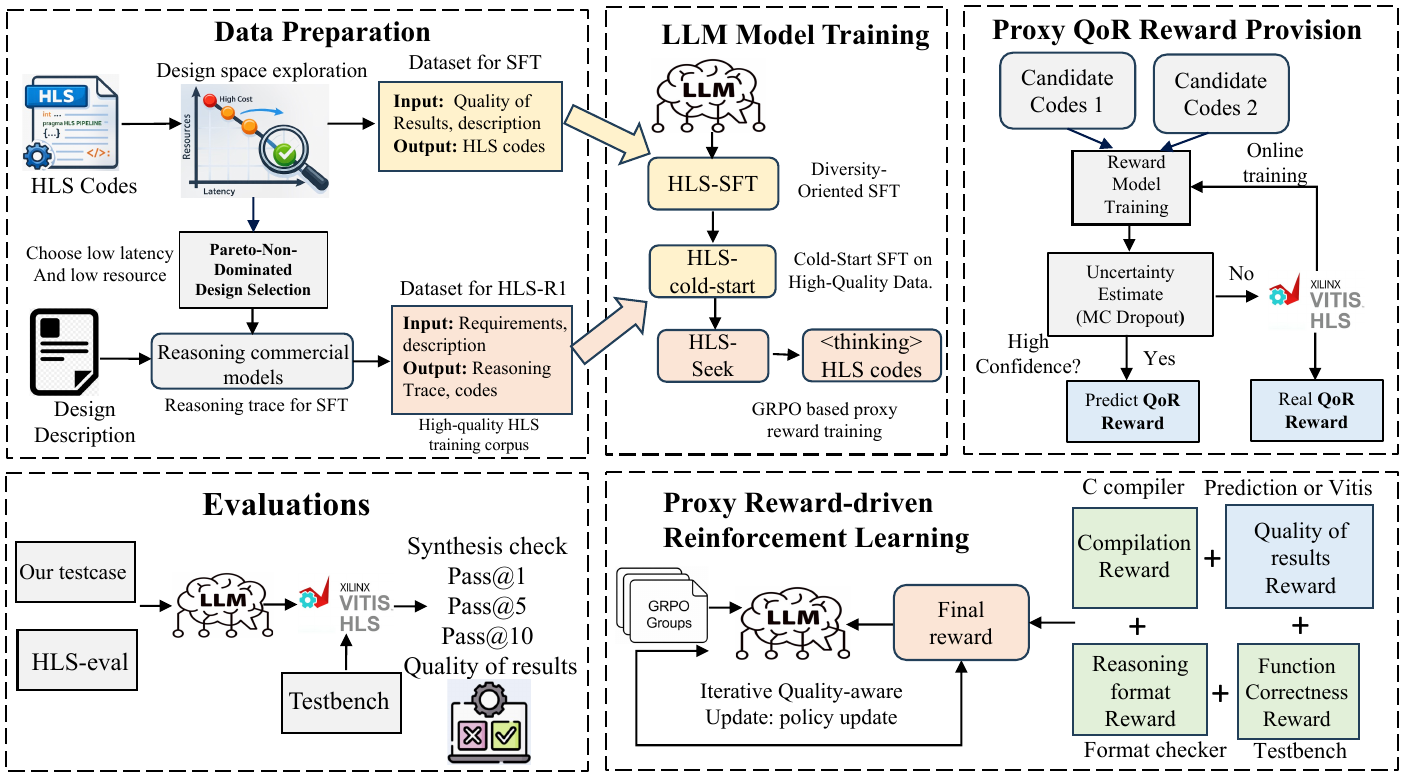}
  \Description{Overview of the HLS-Seek training framework with five components.}
  \caption{Overview of the HLS-Seek framework: data preparation (top-left), three-stage LLM training (top-center), proxy reward with uncertainty-aware switching (top-right), GRPO policy updates (bottom-right), and evaluation (bottom-left).}
  \label{fig:training flow}
\end{figure*}

\vskip -4pt\relax
\section{Background and Motivation}

% \subsection{LLM for RTL Task}
% \td{no need this part at beginning, put it into end of the paper as related work}
% LLM-based RTL generation has progressed from prompt engineering~\cite{blocklove2023chip,thakur2023verigen} and benchmarking~\cite{Liu2023verilogeval} to domain-specific fine-tuning. RTLCoder~\cite{liu2024rtlcodera} demonstrated that SFT with automated data generation can surpass GPT-3.5, while CodeV~\cite{zhao2024codev}, BetterV~\cite{pei2024betterv}, and OriGen~\cite{cui2024origen} further improved quality through data augmentation and self-reflection. Multi-agent frameworks such as MAGE~\cite{zhao2024mage} extended LLMs to iterative RTL design. More recently, attention has shifted toward QoR-aware generation: PPA-RTL~\cite{zhao2025ppartl} applies DPO for PPA-targeted RL training, and ChipSeek~\cite{chen2025chipseek} uses hierarchical reward-driven RL with real synthesis feedback to surpass human-written PPA on RTLLM benchmarks.

% However, both QoR-aware paradigms face limitations when extended to HLS. DPO cannot capture magnitude-sensitive QoR differences arising from small pragma changes---two functionally equivalent HLS programs may differ dramatically in post-synthesis performance. Real synthesis-in-the-loop RL, as employed by ChipSeek, is computationally prohibitive for HLS where each synthesis run takes minutes and the pragma configuration space is exponentially large. These challenges motivate our learned reward proxy approach.

\begin{table}[t]
\centering
\caption{Comparison of LLM-Assisted HLS Code Tasks. NL = Natural Language, AST = Abstract Syntax Tree, RAG = Retrieval-Augmented Generation, HLS-C = HLS-compatible C/C++.}
\label{tab:hls_related}
\footnotesize
\renewcommand{\arraystretch}{1.10}
\renewcommand{\tabularxcolumn}[1]{m{#1}}

\begin{tabularx}{\linewidth}{|>{\centering\arraybackslash}m{1.2cm}
                             |>{\centering\arraybackslash}X
                             |>{\centering\arraybackslash}m{0.8cm}
                             |>{\centering\arraybackslash}X
                             |>{\centering\arraybackslash}X|}
\hline
\textbf{Method} & \textbf{Task} & \textbf{Input} & \textbf{LLM Strategy} & \textbf{Goal} \\ \hline
HLSPilot~\cite{xiong2024hlspilot} & HLS Code \newline Optimization & C/HLS-C & Prompt Engineering & QoR \\ \hline
RALAD~\cite{xu2024optimizing} & HLS Code \newline Optimization & HLS-C & RAG & QoR \\ \hline
C2HLSC~\cite{collini2024c2hlsc} & HLS Code \newline Optimization & C & Prompt Engineering & Functional,\newline QoR \\ \hline
SynthAI~\cite{sheikholeslam2024synthai} & HLS Code \newline Generation & Spec & Multi-Agent & Functional \\ \hline
Gai et al.~\cite{gai2025exploring} & HLS Code \newline Generation & NL & Fine-tuning & Functional \\ \hline
SAGE-HLS~\cite{khan2025sagehls} & HLS Code \newline Generation & NL+AST & Fine-tuning & Functional \\ \hline
\textbf{HLS-Seek} & \textbf{HLS Code \newline Generation} & \textbf{NL} & \textbf{Fine-tuning \newline Uncertainty-aware proxy reward} & \textbf{Functional,\newline QoR} \\ \hline
\end{tabularx}
\end{table}
% \vspace{-0.5em}

% \subsection{LLM for HLS \rv{Code Generation}}
% \td{talk about they are not QoR aware.}
As summarized in Table~\ref{tab:hls_related}, LLM-based HLS research falls into two categories: \textit{HLS code optimization} and \textit{HLS code generation}.

\textbf{HLS Code Optimization.} These methods take existing C source code as input and apply LLMs to pragma insertion or design transformation for improved QoR. HLSPilot~\cite{xiong2024hlspilot} uses prompt engineering to integrate profiling and pipeline optimization; RALAD~\cite{xu2024optimizing} employs RAG to retrieve relevant optimization strategies for pragma insertion; and C2HLSC~\cite{collini2024c2hlsc} iteratively refactors software C into HLS-synthesizable code via prompt engineering. While effective in their target scenarios, these inference-time strategies rely on the availability of correct C source code and do not improve the underlying model's generative capability.

\textbf{HLS Code Generation.} These methods target direct code generation from natural language or high-level specifications. SynthAI~\cite{sheikholeslam2024synthai} applies multi-agent reasoning frameworks to automate the generation pipeline from specifications. Gai et al.~\cite{gai2025exploring} and SAGE-HLS~\cite{khan2025sagehls} fine-tune code LLMs on HLS datasets to produce synthesizable code from natural language descriptions.

However, as Table~\ref{tab:hls_related} shows, existing HLS code generation methods target only functional correctness, while HLS code optimization methods rely on prompt engineering or RAG without learning QoR-aware representations. None of these approaches trains the model with QoR feedback to jointly optimize functional correctness, latency, and resource usage. Our work fills this gap by introducing uncertainty-aware proxy reward-based RL that enables QoR-aware training without synthesis-in-the-loop. Achieving this requires two ingredients that the following section develops in turn: a reward signal that ranks candidate designs by hardware quality at a cost low enough to sit inside an RL loop, and a training pipeline that equips the model with hardware reasoning before that signal is applied.

\vskip -4pt\relax
\section{Methods}
\vskip -1pt\relax
\subsection{Overall Methods}
\vskip -1pt\relax

Figure~\ref{fig:training flow} illustrates the HLS-Seek framework. We first prepare training data via Bayesian DSE \cite{peng2025forgehls} with Pareto-non-dominated selection and reasoning traces from commercial models. The LLM is then trained in three stages: diversity-oriented SFT, cold-start SFT on high-quality Pareto-proximal data, and GRPO~\cite{shao2024deepseekmath} with a proxy reward combining reasoning format, compilation, functional correctness, and predicted QoR (Section~\ref{sec:reward_details}). The QoR reward is provided by a comparative reward model trained on offline synthesis pairs. During GRPO, MC dropout uncertainty estimation triggers selective fallback to real Vitis HLS synthesis for low-confidence candidates, with results fed back to online update the proxy, keeping synthesis overhead below 5\% at convergence.

\vskip -6pt\relax
\subsection{Data Preparation}
\label{sec:data_prep}
\vskip -1pt\relax

\paragraph{Design Space Identification via Clang AST}
Each HLS kernel is parsed with the Clang frontend to extract loop nests (for unrolling and pipelining) and array declarations (for partitioning strategies), constructing a kernel-specific pragma configuration space. Since this space grows exponentially, we use Bayesian optimization for efficient exploration.

\paragraph{HLS Corpus via Bayesian DSE}
Given $\mathcal{P}$, we apply Gaussian process-based Bayesian optimization with Expected Improvement (EI) to efficiently explore the design space \cite{peng2025forgehls}. After $K$ iterations, we obtain evaluated configurations with their synthesis-measured QoR across five dimensions (Latency, LUT, DSP, BRAM, FF). The complete DSE dataset is preserved for diversity-oriented SFT. To facilitate cold-start reasoning training aimed at superior QoR, we isolate Pareto-non-dominated designs, ensuring the model learns from configurations that represent optimal trade-offs across all hardware metrics.

\paragraph{Pareto-Proximal Quality-Diversity Sampling.}
Retaining only the Pareto front $\mathcal{F}$ yields too few training examples per kernel—often as few as 3–5 designs—which limits the LLM's ability to learn fine-grained resource–latency trade-offs. We therefore construct the training set using a two-tier selection strategy. \textit{First}, all Pareto-optimal designs are included. \textit{Second}, for each kernel, we augment with up to $K_{\mathrm{near}}$ near-Pareto designs selected via a two-step process:
(\textit{i})~rank dominated designs by their normalized Euclidean distance to the Pareto front in the 5-dimensional objective space,
\begin{equation}
d(\mathbf{p}) = \min_{\mathbf{p}^* \in \mathcal{F}} \left\| \frac{f(\mathbf{p}) - f(\mathbf{p}^*)}{f_{\max} - f_{\min}} \right\|_2,
\label{eq:pareto_dist}
\end{equation}
where $f_{\max}$ and $f_{\min}$ are the per-metric maximum and minimum across all evaluated designs;
(\textit{ii})~apply greedy farthest-point sampling among the top-ranked candidates (those with $d(\mathbf{p}) < \epsilon$) to maximize diversity in the design space.
This yields a training corpus that balances \textit{quality} (proximity to the Pareto front) and \textit{diversity} (coverage of different resource–latency trade-off regions), providing the LLM with a richer set of competitive design alternatives per kernel.

\paragraph{Reasoning Traces for Cold-Start SFT}
To bootstrap hardware-aware reasoning, we construct a cold-start SFT dataset of \textit{reasoning traces}. For each natural language specification $x$, we query a commercial reasoning model (GPT-o1) to produce a chain-of-thought trace $t$ that articulates step-by-step pragma selection rationale. The resulting dataset $\mathcal{D}_{\mathrm{SFT}} = \{(x_i, t_i, y_i)\}_{i=1}^{N}$ pairs each specification with a reasoning trace and the corresponding HLS implementation, ensuring the base model acquires structured hardware reasoning before RL.

\vskip -6pt\relax
\subsection{LLM Training with Proxy Reward}
\vskip -1pt\relax

\paragraph{Stage 1: Diversity-Oriented SFT}
We first fine-tune the base LLM on a broad HLS corpus where inputs consist of natural language descriptions paired with target QoR specifications, and outputs are the corresponding HLS implementations. This stage prioritizes diversity over optimality by exposing the model to diverse pragmas and coding patterns. It builds a basic understanding of HLS syntax and specification mapping, forming a solid foundation for later QoR-focused training.

\paragraph{Stage 2: Cold-Start SFT on High-Quality Data.}
Starting from the diversity-trained model, we perform a second round of SFT on the high-quality dataset obtained via Pareto-Proximal Quality-Diversity Sampling (Section~\ref{sec:data_prep}). This dataset $\mathcal{D}_{\mathrm{SFT}} = \{(x_i, t_i, y_i)\}_{i=1}^{N}$ pairs each specification with a reasoning trace and a QoR-competitive HLS implementation drawn from or near the Pareto front. Training on this curated corpus sharpens the model's ability to generate hardware-efficient code with structured reasoning traces, producing the SFT model.

\paragraph{Stage 3: GRPO-based RL}
\label{sec:reward_details}
Starting from the SFT model, we apply GRPO~\cite{shao2024deepseekmath}. For each prompt $x$, we sample $G$ candidates $\{a_i\}_{i=1}^G \sim \pi_\theta(\cdot|x)$. Each candidate receives a four-component proxy reward:
\begin{equation}
\begin{split}
r(a_i) &= \lambda_f r_f(a_i) + \lambda_{\mathrm{comp}} r_{\mathrm{comp}}(a_i) \\
        &\quad + \lambda_c r_c(a_i) + \lambda_q r_q(a_i).
\end{split}
\label{eq:reward}
\end{equation}
The four reward components are designed to provide layered feedback from syntactic compliance to hardware quality:

\textbf{Reasoning Format Reward} ($r_f$): Encourages structured outputs of the form:
\begin{center}
\small\texttt{<think>}~$\cdots$~\texttt{</think>~<final\_code>}~$\cdots$~\texttt{</final\_code>}
\end{center}
where \texttt{<think>} contains chain-of-thought reasoning about pragma trade-offs and \texttt{</final\_code>} contains the HLS code. This format promotes explicit internal reasoning, which prior work has shown to improve generation stability~\cite{shao2024deepseekmath}. $r_f = 1$ if the output conforms; $r_f = 0$ otherwise.

\textbf{Compilation Reward} ($r_{\mathrm{comp}}$): Encourages syntactically correct HLS-C. The generated code is passed to a C compiler (GCC); $r_{\mathrm{comp}} = 1$ if compilation succeeds, 0 otherwise. This provides a dense, computationally cheap signal that is independent of functional or synthesis-level evaluation. It also checks for dynamic memory allocations, which cannot be synthesized.

\textbf{Functional Correctness Reward} ($r_c$): Encourages semantic correctness relative to the specification. The compiled code is executed against a predefined testbench, and $r_c$ is defined as a strict binary reward, i.e., $r_c \in \{0, 1\}$. It yields $1$ only if the implementation passes all test cases, and $0$ otherwise, ensuring absolute functional correctness.

\textbf{Proxy QoR Reward} ($r_q$): Encourages hardware efficiency without synthesis-in-the-loop. Only candidates that pass functional testing ($r_c = 1$) are eligible. Among these, we compute $r_q$ via \textit{round-robin pairwise aggregation}: for each pair $(i, j)$ of functionally correct candidates, the comparative reward model (Section~\ref{sec:reward_model}) produces a soft probability $p_{ij} = P(a_i \succ a_j)$ via $M{=}10$ MC dropout forward passes. We accumulate win scores across all duels:
\begin{equation}
r_q(a_i) = \frac{1}{|\mathcal{C}|-1} \sum_{j \in \mathcal{C}, j \neq i} p_{ij},
\label{eq:rq}
\end{equation}
where $\mathcal{C}$ is the set of functionally correct candidates in the group. This yields a scalar $r_q \in [0,1]$ representing each candidate's average win-rate, naturally converting pairwise probabilities into per-candidate QoR scores. If only one candidate passes functional testing, it receives $r_q{=}1.0$.

Within each group, the total rewards are normalized to group-relative advantages $\hat{A}_i = (r(a_i) - \mu_G)/(\sigma_G + \epsilon)$, and the policy is updated via the clipped surrogate loss with Kullback--Leibler (KL) regularization. The resulting model is HLS-Seek.

% \paragraph{Objective-Conditioned Generation.}
% \label{sec:obj_cond}
% To enable controllable QoR optimization without retraining, we introduce \textit{objective-conditioned generation}. An explicit objective token $o \in \mathcal{O} = \{\texttt{latency}, \texttt{lut}, \texttt{balanced}\}$ is appended to each prompt as $x' = [x;\, \texttt{[OBJ]}\, o]$. The QoR reward is adapted accordingly: $\mathcal{M}.\mathrm{rank}_o$ applies objective-weighted scoring $\phi_o(\mathbf{q}) = \sum_{k} w_k^{(o)} q_k$, where $w_k^{(o)}$ up-weights the target metric (e.g., $w_{\mathrm{lat}} = 0.6$ for latency, 0.1 for others). During training, objectives are sampled uniformly per batch; at inference time, the user specifies the desired objective token to steer pragma generation.

\vskip -6pt\relax
\subsection{Comparative Reward Model}
\vskip -1pt\relax
\label{sec:reward_model}

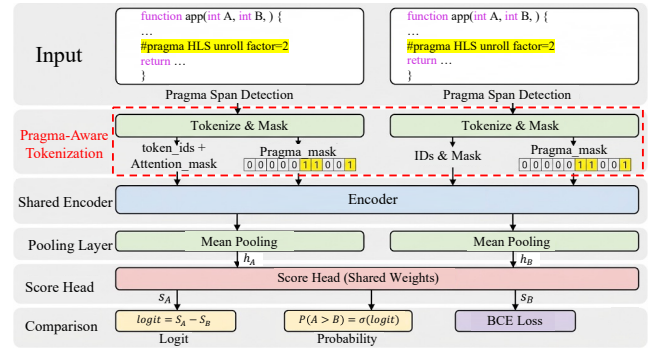
\begin{figure}[t]
  \centering
  \resizebox{0.90\linewidth}{!}{% Figure 3 -- architecture of the comparative reward model, drawn in TikZ so the
% labels are typeset at real document font sizes instead of being shrunk along
% with an exported bitmap/vector figure.
%
% Requires in the preamble:
%   \usepackage{tikz}
%   \usetikzlibrary{arrows.meta,backgrounds}
%
% 85mm wide (one sigconf column) x ~70mm tall. Band padding and arrow lengths
% are deliberately tight so the type can be as large as possible at that height.
\providecommand{\RMPscale}{1.0}

\definecolor{rmpband}{RGB}{255,255,255}
\definecolor{rmpgreen}{RGB}{222,235,214}
\definecolor{rmpblue}{RGB}{198,222,241}
\definecolor{rmppink}{RGB}{247,206,206}
\definecolor{rmpyellow}{RGB}{255,238,180}
\definecolor{rmppurple}{RGB}{219,209,238}
\definecolor{rmphl}{RGB}{255,241,92}
\definecolor{rmpkw}{RGB}{175,0,175}
\definecolor{rmptype}{RGB}{0,0,220}

\begin{tikzpicture}[
  x=\RMPscale mm, y=\RMPscale mm,
  font=\fontsize{8}{8.8}\selectfont,
  box/.style={draw=black, line width=0.4pt, rounded corners=0.4mm,
              align=center, inner sep=1.0mm},
  rowlab/.style={align=center, font=\fontsize{7.6}{8.4}\selectfont},
  ann/.style={align=center, font=\fontsize{6.2}{7.0}\selectfont},
  sym/.style={font=\fontsize{7}{7}\selectfont},
  cmp/.style={font=\fontsize{7}{7.8}\selectfont},
  flow/.style={-{Stealth[length=1.3mm,width=1.3mm]}, line width=0.45pt},
]
\newcommand{\rmpcode}{\ttfamily\fontsize{5.0}{5.9}\selectfont}

% ---------------------------------------------------------------- geometry
\def\Lx{0}   \def\Rx{85}        % picture span
\def\Cx{19}                     % content column starts here
\def\Ac{35}  \def\Bc{69}        % branch centres
\def\bw{30}                     % branch box width
\def\ww{64}                     % shared (full-width) box

% ------------------------------------------------------------ band shading
\begin{scope}[on background layer]
  \fill[rmpband] (\Lx,0)     rectangle (\Rx,-15.2);
  \fill[rmpband] (\Lx,-15.9) rectangle (\Rx,-31.1);
  \fill[rmpband] (\Lx,-31.8) rectangle (\Rx,-40.1);
  \fill[rmpband] (\Lx,-40.8) rectangle (\Rx,-49.1);
  \fill[rmpband] (\Lx,-49.8) rectangle (\Rx,-58.1);
  \fill[rmpband] (\Lx,-58.8) rectangle (\Rx,-69.9);
\end{scope}

% -------------------------------------------------------------- row labels
\node[rowlab]           at (9,-7.6)  {Input};
\node[rowlab, text=red] at (9,-23.5) {Pragma-Aware\\Tokenization};
\node[rowlab]           at (9,-35.9) {Shared\\Encoder};
\node[rowlab]           at (9,-44.9) {Pooling\\Layer};
\node[rowlab]           at (9,-53.9) {Score\\Head};
\node[rowlab]           at (9,-64.3) {Comparison};

% ------------------------------------------------------------- input boxes
\newcommand{\codebox}[1]{%
  \node[box, fill=white, minimum width=\bw mm, anchor=north] at (#1,-0.8) {%
    \rmpcode
    \begin{tabular}{@{}l@{}}
      \textcolor{rmpkw}{function} app(\textcolor{rmptype}{int} A, \textcolor{rmptype}{int} B) \{\\[-0.3ex]
      \hspace{1.2mm}\colorbox{rmphl}{\#pragma HLS unroll factor=2}\\[-0.3ex]
      \hspace{1.2mm}\textcolor{rmpkw}{return} $\cdots$\\[-0.3ex]
      \}
    \end{tabular}};
}
\codebox{\Ac}
\codebox{\Bc}

\node[ann, anchor=north] at (\Ac,-12.1) {Pragma Span Detection};
\node[ann, anchor=north] at (\Bc,-12.1) {Pragma Span Detection};

% --------------------------------------------------------- tokenize layer
\draw[flow] (\Ac,-16.3) -- (\Ac,-18.5);
\draw[flow] (\Bc,-16.3) -- (\Bc,-18.5);
\node[box, fill=rmpgreen, minimum width=\bw mm, anchor=north] at (\Ac,-18.5)
  {Tokenize \& Mask};
\node[box, fill=rmpgreen, minimum width=\bw mm, anchor=north] at (\Bc,-18.5)
  {Tokenize \& Mask};

% pragma mask bit strip
\newcommand{\bitstrip}[2]{%
  \foreach \b [count=\i from 0] in {0,0,0,0,0,1,1,0,0,1} {
    \pgfmathsetmacro{\bx}{#1 - 6.1 + \i*1.35}
    \ifnum\b=1
      \node[draw, line width=0.25pt, inner sep=0pt, fill=rmphl,
            minimum width=1.35mm, minimum height=2.1mm,
            font=\fontsize{4.6}{4.6}\selectfont] at (\bx,#2) {\b};
    \else
      \node[draw, line width=0.25pt, inner sep=0pt, fill=white,
            minimum width=1.35mm, minimum height=2.1mm,
            font=\fontsize{4.6}{4.6}\selectfont] at (\bx,#2) {\b};
    \fi
  }
}

\draw[flow] (\Ac-7.5,-23.6) -- (\Ac-7.5,-25.3);
\draw[flow] (\Ac+7.5,-23.6) -- (\Ac+7.5,-25.3);
\draw[flow] (\Bc-7.5,-23.6) -- (\Bc-7.5,-25.3);
\draw[flow] (\Bc+7.5,-23.6) -- (\Bc+7.5,-25.3);

\node[ann, anchor=north] at (\Ac-7.5,-25.4) {token\_ids +\\Attention\_mask};
\node[ann, anchor=north] at (\Ac+7.5,-25.4) {Pragma\_mask};
\bitstrip{\Ac+7.5}{-29.4}

\node[ann, anchor=north] at (\Bc-7.5,-25.4) {IDs \& Mask};
\node[ann, anchor=north] at (\Bc+7.5,-25.4) {Pragma\_mask};
\bitstrip{\Bc+7.5}{-29.4}

\draw[red, dashed, line width=0.5pt, rounded corners=0.8mm]
  (\Cx+0.6,-17.9) rectangle (\Rx-0.6,-30.8);

% ---------------------------------------------------------------- encoder
\draw[flow] (\Ac-7.5,-31.0) -- (\Ac-7.5,-34.4);
\draw[flow] (\Ac+7.5,-31.0) -- (\Ac+7.5,-34.4);
\draw[flow] (\Bc-7.5,-31.0) -- (\Bc-7.5,-34.4);
\draw[flow] (\Bc+7.5,-31.0) -- (\Bc+7.5,-34.4);
\node[box, fill=rmpblue, minimum width=\ww mm, anchor=north] at (52,-34.4)
  {Encoder};

% ---------------------------------------------------------------- pooling
\draw[flow] (\Ac,-39.5) -- (\Ac,-43.4);
\draw[flow] (\Bc,-39.5) -- (\Bc,-43.4);
\node[box, fill=rmpgreen, minimum width=\bw mm, anchor=north] at (\Ac,-43.4)
  {Mean Pooling};
\node[box, fill=rmpgreen, minimum width=\bw mm, anchor=north] at (\Bc,-43.4)
  {Mean Pooling};

% ------------------------------------------------------------- score head
\draw[flow] (\Ac,-48.5) -- (\Ac,-52.4) node[sym, midway, right=0.4mm] {$h_A$};
\draw[flow] (\Bc,-48.5) -- (\Bc,-52.4) node[sym, midway, right=0.4mm] {$h_B$};
\node[box, fill=rmppink, minimum width=\ww mm, anchor=north] at (52,-52.4)
  {Score Head (Shared Weights)};

% ------------------------------------------------------------- comparison
\draw[flow] (30,-57.5) -- (30,-61.4) node[sym, midway, left=0.4mm]  {$s_A$};
\draw[flow] (52.5,-57.5) -- (52.5,-61.4);
\draw[flow] (74,-57.5) -- (74,-61.4) node[sym, midway, right=0.4mm] {$s_B$};

\node[box, cmp, fill=rmpyellow, anchor=north] at (30,-61.4)
  {$logit = s_A - s_B$};
\node[box, cmp, fill=rmpyellow, anchor=north] at (52.5,-61.4)
  {$P(A{>}B) = \sigma(logit)$};
\node[box, cmp, fill=rmppurple, anchor=north] at (74,-61.4)
  {BCE Loss};

\node[ann, anchor=north] at (30,-66.6) {Logit};
\node[ann, anchor=north] at (52.5,-66.6) {Probability};

\end{tikzpicture}}
  \Description{Architecture of the comparative reward model.}
  \caption{Architecture of the comparative reward model.}
  \label{fig:reward_model_arch}
\end{figure}

Unlike prior HLS QoR prediction models that regress absolute values, our model performs pairwise comparison, which is simpler and directly aligned with GRPO's relative advantage estimation. We adopt the Bradley-Terry~\cite{bradley1952rank} Siamese architecture (Figure~\ref{fig:reward_model_arch}), widely used in reinforcement learning from human feedback (RLHF) reward modeling~\cite{ouyang2022training}, as it guarantees ranking transitivity and enables fast inference with a lightweight encoder.

To help the model focus on the QoR-critical parts of HLS code, we introduce \textit{Pragma-Aware Tokenization}: before encoding, we identify all \texttt{\#pragma HLS} directives in the source code and prepend a special \texttt{[PRAGMA]} token to each pragma line. This allows the encoder's attention to naturally upweight pragma regions---which govern unrolling, pipelining, and array partitioning---without modifying the underlying Transformer architecture. Each HLS design, represented as the concatenation of its functional description and pragma-annotated code, is encoded by a shared Transformer encoder with mean pooling (768-d), then mapped to a scalar score by a shared two-layer multilayer perceptron (MLP), 768$\to$384$\to$1. All weights are shared between branches, ensuring ranking transitivity. Three dropout layers (rate 0.2) are retained active at inference for MC Dropout uncertainty estimation. For each application, different HLS implementations from DSE are paired and assigned a three-level graded target. Throughout this paper, design $i$ \emph{Pareto-dominates} design $j$ if $i$ is no worse than $j$ on all five QoR metrics and strictly better on at least one:
\begin{equation}
y_{ij} = \begin{cases}
1.0 & \text{if } i \text{ Pareto-dominates } j, \\
0.5 & \text{if } i \text{ does not dominate } j \text{ but has lower latency,} \\
0.0 & \text{otherwise.}
\end{cases}
\label{eq:label}
\end{equation}
Full Pareto dominance ($y_{ij}{=}1$) captures designs that are no worse on every dimension and strictly better on at least one, while latency-only superiority ($y_{ij}{=}0.5$) provides partial credit for the common case where designs trade off latency against resource usage. True ties and pairs with relative gap below a threshold $\delta$ are discarded to reduce label noise. Each design is mapped to a scalar score $s_i$ via the reward model's score head. We define the score difference $d_{ij}$ and its corresponding graded comparison score $\hat{y}_{ij}$ as
\begin{equation}
\begin{split}
d_{ij} &= s_i-s_j, \\
\hat{y}_{ij} &= \sigma(d_{ij}), \\
\mathcal{L}_{\mathrm{RM}} &= \mathcal{L}_{\mathrm{BCEWithLogits}}(d_{ij}, y_{ij}).
\end{split}
\label{eq:rm_loss}
\end{equation}
Here $\mathcal{L}_{\mathrm{BCEWithLogits}}$ denotes the binary cross-entropy (BCE) loss computed on logits, and $y_{ij}{=}0.5$ is a soft target representing partial credit rather than a categorical class or a calibrated binary preference probability. During GRPO, only functionally correct candidates compete for QoR reranking via group-wise pairwise comparison to assign $r_q$. This graded scheme provides a learning signal even when full Pareto dominance is rare: latency-superior designs receive partial credit ($y_{ij}{=}0.5$), while true ties are excluded from reward-model training.

\paragraph{Uncertainty-Aware Proxy Reward Switching.}
\label{sec:uncertainty}
A key risk of proxy reward models in online RL is \textit{reward hacking}: as GRPO explores the pragma space, the policy may produce designs that exploit the reward model's out-of-distribution blind spots. To mitigate this risk, we equip the reward model (dropout rate 0.2) with \textit{Monte Carlo dropout uncertainty estimation}~\cite{gal2016dropout}. We retain dropout layers active during inference and perform $M{=}10$ stochastic forward passes per candidate:
\begin{equation}
u_i = \mathrm{Var}_{m=1}^{M}\bigl[s^{(m)}(a_i)\bigr],
\end{equation}
where $s^{(m)}(a_i)$ is the score head output under the $m$-th dropout mask. When $u_i > \tau_u{=}0.1$ for any candidate in the group, we invoke real Vitis HLS synthesis for that candidate and use the result to compute $r_q$. This synthesis result is added to a dynamic replay buffer $\mathcal{B}$, which is used to fine-tune $\mathcal{M}$ online every $K_{\mathrm{update}}{=}100$ steps (learning rate $2{\times}10^{-6}$, 50 fine-tuning steps per update). This adaptive mechanism serves two purposes: (\textit{i}) it grounds rewards at policy frontier states where the proxy is least reliable, reducing reward-hacking risk; and (\textit{ii}) it continuously improves $\mathcal{M}$'s coverage as the policy explores new regions. Thus, MC dropout estimates uncertainty and routes high-uncertainty candidates to real synthesis, mitigating reward errors when proxy coverage is limited.

\vskip -4pt\relax
\section{Experiments}
\vskip -1pt\relax
\subsection{Experimental Setup}
\vskip -1pt\relax

\textbf{Model and Hardware.}
All experiments use \textbf{Qwen2.5-Coder-7B-Instruct}~\cite{hui2024qwen2} as the base model, fine-tuned on 2$\times$ NVIDIA H200 GPUs with bfloat16 (bf16) precision and the AdamW optimizer. HLS synthesis and QoR evaluation use \textbf{Vitis HLS 2023.1} targeting an AMD Alveo U280 FPGA (xcu280-fsvh2892-2L-e, Virtex UltraScale+ with high-bandwidth memory (HBM)) with a clock period of 10\,ns (100\,MHz). Frontier model baselines (GPT-5.1, DeepSeek-V3.2, Gemini-3-Pro, Qwen3-235B) are accessed via the ChatFire API\footnote{\url{https://api.chatfire.cn/}} with temperature 0.7 and default decoding settings.

\begin{figure}[t]
  \centering
  \begin{minipage}{0.48\linewidth}
    \centering
    \includegraphics[width=0.90\linewidth]{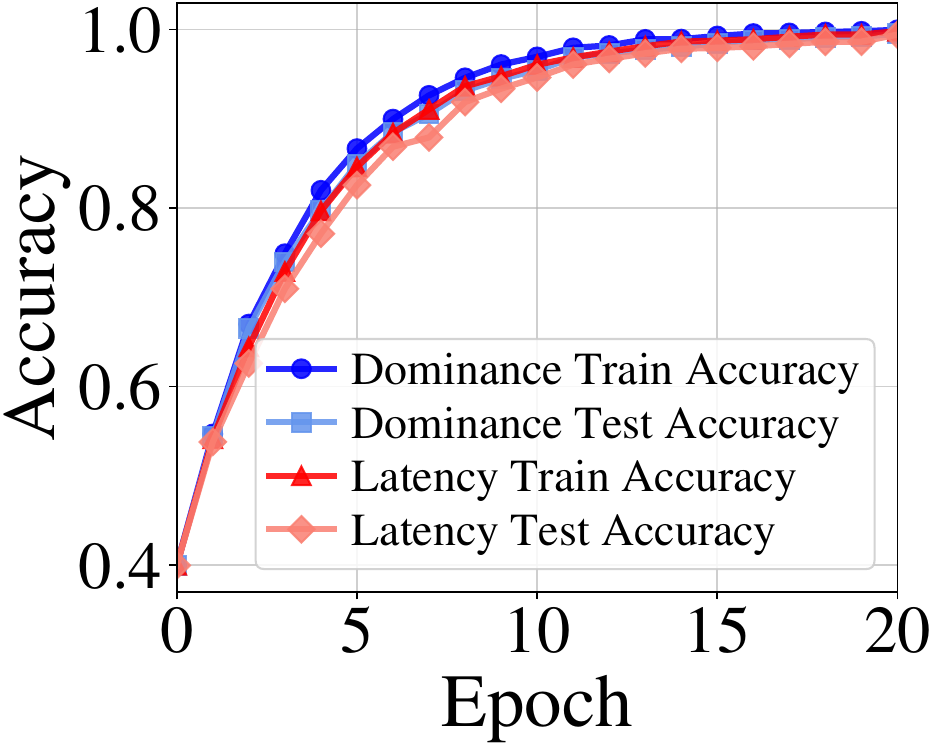}
    \subcaption{Reward model accuracy}
  \end{minipage}
  \hfill
  \begin{minipage}{0.48\linewidth}
    \centering
    \includegraphics[width=0.90\linewidth]{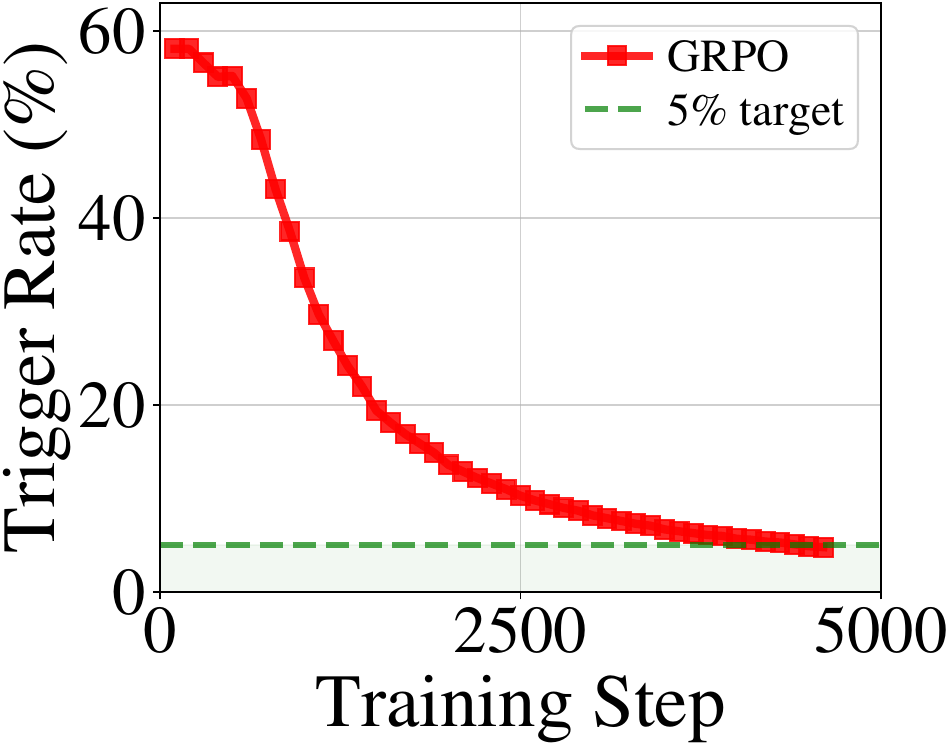}
    \subcaption{MC dropout trigger rate}
  \end{minipage}
  \Description{(a) Reward model accuracy curve. (b) MC dropout trigger rate curve.}
  \caption{Evaluation of the proxy reward mechanism.}
  \label{fig:ranker_accuracy}
\end{figure}

\textbf{Dataset.}
Training data is aggregated from three open-source HLS datasets: ForgeHLS~\cite{peng2025forgehls}, SAGE-HLS~\cite{khan2025sagehls}, and HLStrans~\cite{hlstrans2025}, comprising a total of 12,661 applications and 450,273 samples. The cold-start SFT (Stage 2) selects 418 applications and applies our Pareto-Proximal Quality-Diversity Sampling to curate high-quality samples paired with reasoning traces, and GRPO training (Stage 3) uses a further 1,013 applications. The diversity-oriented SFT (Stage 1) uses the remaining 11,230 applications with QoR descriptions for broad pragma coverage. The comparative reward model is trained on pairwise examples constructed from DSE-generated HLS variants of the 1,013 applications used for GRPO (Section~\ref{sec:proxy_effects}). Evaluation is conducted on the independent HLS-Eval benchmark (43 kernels) and a randomly selected ForgeHLS test set (108 applications), with no overlap between training and evaluation data.

\textbf{SFT Training.}
Both SFT stages apply low-rank adaptation (LoRA) fine-tuning with rank $r{=}64$, scaling $\alpha{=}128$, and dropout 0.05, targeting the projection and feed-forward network (FFN) modules in transformer block. Each stage is trained for 3 epochs with a learning rate of $2{\times}10^{-4}$, batch size 16, and gradient checkpointing enabled.

\begin{table*}[t]
\caption{Evaluation results on ForgeHLS test set (108 apps) and the independent HLS-Eval benchmark (Higher is better). }
\label{function_compare}
\centering
\renewcommand{\arraystretch}{0.95}
\small
\newcolumntype{Y}{>{\centering\arraybackslash}X}
\begin{tabularx}{\textwidth}{@{}>{\raggedright}p{1.8cm} l r *{6}{Y} *{4}{Y}@{}}
\toprule
\textbf{Type} & \textbf{Model} & \textbf{Size} &
\multicolumn{6}{c}{\textbf{ForgeHLS Test Set (\%)}} &
\multicolumn{4}{c}{\textbf{HLS-Eval (\%)}} \\
\cmidrule(lr){4-9} \cmidrule(lr){10-13}
 & & &
\multicolumn{3}{c}{\textbf{Syntax Correctness}} &
\multicolumn{3}{c}{\textbf{Functional Correctness}} &
\multicolumn{2}{c}{\textbf{Syntax Corr.}} &
\multicolumn{2}{c}{\textbf{Func. Corr.}} \\
\cmidrule(lr){4-6} \cmidrule(lr){7-9} \cmidrule(lr){10-11} \cmidrule(lr){12-13}
 & & &
\textbf{pass@1} & \textbf{pass@5} & \textbf{pass@10} &
\textbf{pass@1} & \textbf{pass@5} & \textbf{pass@10} &
\textbf{pass@1} & \textbf{pass@5} &
\textbf{pass@1} & \textbf{pass@5} \\
\midrule

% Foundational Models
\multirow{4}{*}{\parbox{1.8cm}{Foundational\\Models}}
 & GPT-5.1 & - & \textbf{84.3} & \underline{89.8} & \underline{91.7} & \textbf{73.1} & \underline{80.6} & \underline{82.4} & \textbf{91.7} & \textbf{97.7} & \textbf{71.7} & \underline{79.1} \\
 & DeepSeek-V3.2 & - & 73.1 & 88.0 & 89.8 & 62.0 & 77.8 & 80.6 & 77.3 & 93.0 & 51.9 & 65.1 \\
 & Gemini-3-Pro & - & 62.0 & 88.0 & 90.7 & 55.6 & 78.7 & 80.6 & 67.1 & \underline{95.3} & 55.9 & \underline{79.1} \\
 & Qwen3-235B & 235B & 33.3 & 42.6 & 43.5 & 24.1 & 37.0 & 39.8 & 61.8 & 79.1 & 43.9 & 67.4 \\
\midrule

% Code Models
\multirow{6}{*}{Code Models}
 & Qwen2.5-Coder & 32B & 75.0 & 86.1 & 87.0 & 61.1 & 75.9 & 75.9 & 77.0 & 88.4 & 46.8 & 58.1 \\
 & Qwen2.5-Coder & 14B & 77.8 & 86.1 & 87.0 & 63.9 & 76.9 & 77.8 & 73.6 & 81.4 & 36.7 & 44.2 \\
 & DeepSeek-Coder & 6.7B & 63.9 & 86.1 & 88.0 & 46.3 & 74.1 & 78.7 & 69.0 & 93.0 & 29.1 & 46.5 \\
 & Qwen2.5-Coder & 7B & 41.7 & 76.9 & 82.4 & 24.1 & 63.0 & 69.4 & 45.4 & 83.7 & 16.0 & 41.9 \\
 & CodeLlama & 7B & 17.6 & 57.4 & 75.0 & 10.2 & 38.0 & 52.8 & 19.3 & 62.8 & 10.6 & 39.5 \\
 & Starcoder2 & 15B & 19.4 & 31.5 & 49.1 & 8.1 & 13.9 & 24.1 & 8.6 & 41.9 & 4.8 & 34.9 \\
\midrule

% HLS-specific models
\multirow{1}{*}{Gai et al.~\cite{gai2025exploring}}
 & CodeLlama & 7B & 22.2 & 28.7 & 45.4 & 15.7 & 17.6 & 27.8 & 6.3 & 48.8 & 1.4 & 27.9 \\
\midrule

\multirow{1}{*}{SAGE-HLS~\cite{khan2025sagehls}}
 & Qwen2.5-Coder & 7B & 77.8 & 88.0 & 88.0 & 63.9 & 76.9 & 77.8 & 67.8 & 76.7 & 34.8 & 41.9 \\
\midrule

% Ours
\multirow{2}{*}{\textbf{Ours}}
 & HLS-Seek (SFT) & 7B & 78.7 & 89.8 & 90.7 & 66.7 & 79.6 & 82.4 & 77.5 & 88.4 & 50.7 & 60.5 \\
 & HLS-Seek (GRPO) & 7B & \underline{81.5} & \textbf{91.7} & \textbf{93.5} & \underline{71.3} & \textbf{84.3} & \textbf{87.0} & \underline{84.7} & \underline{95.3} & \underline{68.9} & \textbf{81.4} \\
\bottomrule
\end{tabularx}
\end{table*}

\textbf{GRPO Training.}
Starting from the SFT checkpoint, GRPO uses group size $G{=}4$, learning rate $5{\times}10^{-7}$, clip ratio $\epsilon{=}0.2$, KL coefficient $\beta{=}0.02$, and 2 proximal policy optimization (PPO) epochs per batch. Generation employs temperature 0.9, top-$p$ 0.95, and a maximum of 2,048 new tokens.

\textbf{Comparative Reward Model.}
The pairwise reward model is built on \textbf{UniXcoder-base}~\cite{guo2022unixcoder}, fine-tuned for 20 epochs with learning rate $1{\times}10^{-5}$, batch size 16, and dropout 0.2. The maximum input length is set to cover the full code length of all kernels in the dataset. Pairs with relative QoR difference below 10\% and true ties are discarded to reduce label noise. The model is optimized using the graded BCE-with-logits objective in Equation~\ref{eq:rm_loss}.

\vskip -6pt\relax
\subsection{Evaluation Metrics and Benchmarks}
\vskip -1pt\relax

\textbf{Syntax Correctness.}
We measure \emph{syntax correctness} as the compilation pass rate—the fraction of generated HLS programs that successfully pass Vitis HLS compilation without syntax or type errors. A design that fails to compile cannot be synthesized or tested, so this metric captures the fundamental quality bar for HLS code generation.

\textbf{Functional Correctness.}
We report functional correctness using the pass@$k$ metric~\cite{chen2021evaluating}, which estimates the probability that at least one correct solution appears in $k$ independent samples. Correctness is determined by running the generated HLS code against provided testbenches. We report pass@1, pass@5, and pass@10.

\textbf{QoR-Aware Evaluation.}
Beyond functional correctness, we evaluate whether LLMs can generate HLS code that is also hardware-efficient. We conduct a QoR-aware experiment on benchmark by appending the optimization directive ``\texttt{Write HLS code for least latency with least resource}'' to each prompt, instructing the model to simultaneously minimize execution latency (cycles) and resource usage (LUT, DSP, BRAM, FF). We sample $N$ candidates, filter for functional correctness, synthesize all passing designs with Vitis HLS, and compare the resulting QoR against reference designs using Pareto dominance.

\textbf{Data Independence.}
To ensure no data leakage between training and evaluation, we verify separation via MD5 hash deduplication on every training and evaluation record, using both raw hashes and a normalized form (comments and whitespace stripped). Across all train-eval splits (SFT/GRPO training vs.\ ForgeHLS test set, and vs.\ HLS-Eval), we observe zero exact and zero normalized matches.

\vskip -6pt\relax
\subsection{Main Results}
\vskip -1pt\relax

\subsubsection{Proxy Reward Effects}
\label{sec:proxy_effects}
\vskip -0.5pt\relax

% \begin{figure}[t]
%   \centering
%   \includegraphics[width=\linewidth]{figs/reward_model.pdf}
%   \caption{Pairwise ranking accuracy of the comparative reward model before and after fine-tuning. Training on DSE-generated HLS pairs improves accuracy from 39.5\% to 99.75\%.}
%   \label{fig:reward_model}
% \end{figure}

The proxy reward model only needs to serve GRPO training, which covers 1,013 applications. It is trained on 398,734 pairwise samples constructed from DSE-generated HLS variants of these applications. Crucially, the task is \textit{graded pairwise comparison}---the model assigns full credit to Pareto dominance (target 1.0) and partial credit to latency-only superiority (target 0.5), rather than predicting absolute resource or latency values. This relative comparison is inherently easier than regression, as the three-level targets provide clear margins between quality levels. Figure~\ref{fig:ranker_accuracy}a reports accuracy separately on the Pareto-dominance and latency-only subsets; both exceed 99\% by epoch 20 with minimal train-test gaps. The final test accuracy reaches \textbf{99.53\%} on Pareto-dominance pairs and \textbf{99.42\%} on latency-only pairs, confirming that both levels of the graded reward signal are reliably predicted.

Figure~\ref{fig:ranker_accuracy}b shows the MC dropout uncertainty trigger rate during GRPO training. Early in training, over 55\% of candidates trigger real Vitis HLS synthesis due to high reward model uncertainty on the policy's out-of-distribution outputs. As training progresses and the reward model is updated with synthesis results, the trigger rate decreases monotonically, dropping below the 5\% target after approximately 4,500 steps. This confirms that the uncertainty-aware switching mechanism is self-correcting: it grounds rewards when most needed (early training) while adding negligible overhead at convergence.

\begin{table*}[t]
\centering
\scriptsize
\renewcommand{\arraystretch}{0.72}
\caption{QoR comparison on all functionally correct kernels. ``--'' indicates that synthesis exceeded the 30-minute timeout. Bold = lowest latency or Pareto-dominant entry.}
\newcolumntype{Y}{>{\centering\arraybackslash}X}
\resizebox{\textwidth}{!}{%
\begin{tabularx}{\textwidth}{|l|Y|Y|Y|Y|}
\hline
\multirow{2}{*}{Kernel} & SAGE-HLS~\cite{khan2025sagehls} & Gai et al.~\cite{gai2025exploring} & GPT-5.1 & \textbf{HLS-Seek} \\
 & (Lat./DSP/FF/BRAM/LUT) & (Lat./DSP/FF/BRAM/LUT) & (Lat./DSP/FF/BRAM/LUT) & (Lat./DSP/FF/BRAM/LUT) \\
\hline
fgnn\_linear       & 554/5/3305/1/4791 & -- & 169/12/5730/0/2864 & \textbf{116}/648/90762/0/67419 \\
\hline
block\_freq        & 529/11/1290/0/1289 & 529/11/1290/0/1289 & \textbf{161}/11/3419/4/4717 & \textbf{161}/11/3419/4/4717 \\
\hline
global\_add\_pool   & 18/2/3150/2/2533 & 18/2/404/0/660 & 1626/2/606/0/1164 & \textbf{10}/16/8816/0/5906 \\
\hline
gemver             & 8252/11/5399/4/5285 & 2617/33/27898/0/8646 & 8042/22/2967/0/3202 & \textbf{1938}/44/52621/0/17716 \\
\hline
seidel\_2d         & 758521/3/1222/0/1732 & \textbf{61073}/24/17611/0/11468 & 758521/3/1222/0/1732 & 1099761/3/829/0/1367 \\
\hline
spam\_dotProduct   & 1152/64/1409/0/5746 & 1026/2/46/0/171 & \textbf{36}/64/1042/0/3705 & 132/16/1782/0/2660 \\
\hline
max\_widen\_port   & \textbf{81}/0/1158/2/1235 & 130/0/17/0/109 & \textbf{81}/0/1787/4/2001 & 129/0/19/0/117 \\
\hline
data\_forwarding  & 1024/0/1025/0/2715 & 1026/0/25/0/77 & 1040/0/1889/4/1962 & \textbf{514/0/309/0/638} \\
\hline
hamming\_encoder   & \textbf{512}/0/513/0/31139 & 1026/0/25/0/99 & 1026/0/25/0/105 & 514/0/23/0/129 \\
\hline
instr\_pipeline    & 1023/0/1024/0/45344 & 1026/0/25/0/116 & 1028/0/29/0/116 & \textbf{641/0/118/0/810} \\
\hline
mag\_comparator    & \textbf{1026}/0/303/4/402 & 2050/0/25/0/117 & \textbf{1026/0/25/0/116} & \textbf{1026}/0/25/0/190 \\
\hline
comparator\_8bit  & 2063/0/12356/2/22019 & 1026/0/25/0/115 & 1041/0/2555/6/3089 & \textbf{514/0/25/0/450} \\
\hline
mux\_4\_to\_1      & -- & 2074/0/2383/2/2451 & 1027/0/42/0/103 & \textbf{515/0/64/0/1296} \\
\hline
nand\_gate         & \textbf{512}/0/513/0/28333 & 3086/0/112/0/34639 & 1026/0/25/0/81 & \textbf{514/0/23/0/85} \\
\hline
pll\_2x            & 1026/0/24/0/120 & -- & 1028/0/29/0/77 & \textbf{514}/0/36/0/213 \\
\hline
pll\_8x            & \textbf{512}/0/513/0/18851 & 2050/0/25/0/78 & 1026/0/25/0/77 & 1025/0/36/0/85 \\
\hline
xor\_gate          & 512/0/513/0/26285 & 2050/0/372/0/486 & \textbf{4}/0/15/0/1082 & \textbf{1026/0/27/0/131} \\
\hline
karaoke\_proc      & 1045/8/1833/2/1756 & -- & 1031/8/801/0/668 & \textbf{263}/32/2288/0/1879 \\
\hline
video\_proc        & 1048578/0/262/8/425 & 1048578/0/66/0/245 & 1048580/0/70/0/245 & \textbf{262401}/0/19515/0/63162 \\
\hline
binary\_counter    & -- & 1030/0/851/0/929 & \textbf{512}/0/513/0/335267 & 518/0/1666/0/1810 \\
\hline
hs\_6bit\_adder   & \textbf{512}/0/513/0/18877 & -- & 1041/0/2526/6/3047 & \textbf{514/0/36/0/239} \\
\hline
ittyBittyComp      & 1026/0/24/0/98 & 2060/0/2060/5/2501 & 1026/0/91/0/216 & \textbf{258/0/343/0/1187} \\
\hline
periph\_interface  & \textbf{1026}/0/61/0/165 & \textbf{1026/0/25/0/116} & \textbf{1026/0/25/0/116} & \textbf{1026}/0/25/0/136 \\
\hline
video\_ctrl        & -- & -- & \textbf{2073606}/0/220/0/322 & \textbf{2073606}/0/220/0/322 \\
\hline
crc32              & \textbf{1027}/0/291/2/2342 & -- & \textbf{1027}/0/81/0/2211 & 11265/0/64/0/428 \\
\hline
digital\_phase     & 1051/42/2943/0/7452 & 1060/42/3023/0/7535 & \textbf{1049}/32/2321/0/9060 & \textbf{1049}/32/2321/0/9060 \\
\hline
hazard\_detect     & -- & -- & 1026/0/25/0/297 & \textbf{514}/0/437/0/3717 \\
\hline
portfolio\_opt     & -- & -- & \textbf{23408932/11/2053/0/2479} & 318983060/19/2771/4/2817 \\
\hline
vector\_add        & 1026/0/301/6/366 & -- & 1026/0/25/0/116 & \textbf{514}/0/369/0/679 \\
\hline
crc8              & 1026/0/65/0/274 & -- & 1026/0/22/0/236 & \textbf{641}/0/66/0/1240 \\
\hline
\multicolumn{5}{|c|}{\textit{\textbf{Lowest latency}: HLS-Seek \textbf{19}/30, GPT-5.1 11/30, SAGE-HLS 8/30, Gai et al.\ 2/30}} \\
\hline
\multicolumn{5}{|c|}{\textit{\textbf{Pareto Dominance}.\ (HLS-Seek wins/loses): vs SAGE-HLS \textbf{5}/0, vs Gai \textbf{4}/1, vs GPT-5.1 3/3}} \\
\hline
\end{tabularx}%
}
\label{tab:qor_compare}
\end{table*}

\subsubsection{HLS Code Generation Results}
\vskip -0.5pt\relax

Table~\ref{function_compare} reports syntax correctness and functional correctness across all baselines. \textbf{HLS-Seek} achieves the strongest results of any 7B model in the table, and on the ForgeHLS test set it leads every baseline at pass@5 and pass@10, including models orders of magnitude larger.

\textbf{Syntax Correctness.}
HLS-Seek achieves \textbf{81.5\%} pass@1 syntax correctness, surpassing all code-specialist baselines despite only 7B parameters. At pass@5 and pass@10 it reaches \textbf{91.7\%} and \textbf{93.5\%}, the highest values in the table, indicating the model rarely fails to produce a compilable design within a small sample budget. Among foundational models, GPT-5.1 (84.3\%) achieves higher pass@1 but falls behind at pass@5 (89.8\%) and pass@10 (91.7\%). Compared with the HLS-specific baseline SAGE-HLS (77.8\%), HLS-Seek achieves a +3.7 pp gain at pass@1 and +3.7 pp at pass@5, confirming the benefit of QoR-aware reinforcement learning.

\textbf{Functional Correctness.}
HLS-Seek achieves \textbf{71.3\%} functional pass@1, and its advantage becomes decisive at higher sample budgets: \textbf{84.3\%} pass@5 and \textbf{87.0\%} pass@10, the highest values in the table, outperforming GPT-5.1 by +3.7 pp and +4.6 pp respectively. Among code models, Qwen2.5-Coder-14B (63.9\%) and DeepSeek-Coder-6.7B (46.3\%) achieve competitive pass@1 but lag significantly at higher budgets. On ForgeHLS, general code models such as Starcoder2-15B achieve low syntax correctness (19.4\% pass@1) and low functional correctness (8.1\% pass@1), illustrating that HLS-specific training is necessary for semantically correct kernel generation.

\textbf{HLS-Eval Benchmark.}
On the cross-benchmark HLS-Eval task, HLS-Seek achieves \textbf{84.7\%} Syntax pass@1 / \textbf{95.3\%} pass@5 and \textbf{68.9\%} Func pass@1 / \textbf{81.4\%} pass@5. While GPT-5.1 achieves higher Syntax pass@5 (97.7\%), HLS-Seek surpasses it on Func pass@5 by +2.3 pp (81.4\% vs.\ 79.1\%). Compared with SAGE-HLS (67.8\% / 76.7\% Syntax, 34.8\% / 41.9\% Func), HLS-Seek achieves substantial gains across all metrics (+16.9 pp Syntax pass@1, +18.6 pp Syntax pass@5, +34.1 pp Func pass@1, +39.5 pp Func pass@5), confirming the effectiveness of proxy reward-based reinforcement learning.

\vskip -4pt\relax
\subsubsection{Quality of Results}
\vskip -0.5pt\relax

Table~\ref{tab:qor_compare} reports post-synthesis QoR (Latency/DSP/FF/BRAM/LUT) for kernels where all four models produce functionally correct and synthesizable designs. We select these kernels from both the HLS-Eval benchmark and our ForgeHLS test set (108 applications). Among commercial models, we choose GPT-5.1 as the representative baseline because it achieves the highest functional correctness in Table~\ref{function_compare}. The two HLS-specific baselines are SAGE-HLS~\cite{khan2025sagehls} and Gai et al.~\cite{gai2025exploring}. All models are prompted with ``\texttt{Write HLS code for least latency with least resource}'' to elicit QoR-optimized outputs. For each kernel, we sample 5 candidates per model and report the QoR of the best functionally correct design (best-of-5), consistent with the pass@5 protocol in Table~\ref{function_compare}.

\textbf{Latency.} HLS-Seek achieves the lowest latency on the majority of kernels. On compute-intensive designs such as \texttt{gemver} (1938 vs.\ 8042 for GPT-5.1), \texttt{video\_proc} (262401 vs.\ 1048580), and \texttt{karaoke\_\allowbreak proc} (263 vs.\ 1031), HLS-Seek reduces latency by 1.4--4.3$\times$ compared to all baselines. This advantage stems from the QoR-aware GRPO training, which rewards aggressive pragma strategies (e.g., loop unrolling, pipelining) that reduce execution cycles.

\textbf{Resource Trade-off.} The latency reduction often comes at the cost of higher resource usage. For example, on \texttt{fgnn\_linear}, HLS-Seek achieves the lowest latency (116) but uses significantly more FF and LUT than GPT-5.1. Conversely, on simpler kernels such as \texttt{comparator\_8bit} and \texttt{data\_forwarding}, HLS-Seek achieves both lower latency and moderate resource consumption, demonstrating that the model can identify efficient pragma configurations when the design space permits.

\textbf{Pareto Dominance.} Considering all five QoR dimensions simultaneously, HLS-Seek Pareto-dominates SAGE-HLS on 5 kernels with 0 reverse, Gai et al.~\cite{gai2025exploring} on 4 kernels with 1 reverse, and GPT-5.1 on 3 kernels with 3 reverse. The clear advantage over HLS-specific baselines (9:1 aggregate) confirms that QoR-aware RL training produces designs that are not merely faster but holistically better across resource dimensions. Against GPT-5.1, the symmetric 3:3 result reflects a natural trade-off: HLS-Seek favors latency-optimized designs while GPT-5.1 tends toward resource-conservative configurations.

\textbf{Comparison with Baselines.} Gai et al.~\cite{gai2025exploring} frequently exceed the 30-minute synthesis timeout, reflecting the limited pragma awareness of its CodeLlama-based fine-tuning. SAGE-HLS produces conservative designs with moderate latency but occasionally incurs extremely high LUT usage (e.g., \texttt{nand\_gate}: 28333 LUT, \texttt{hamming\_\allowbreak encoder}: 31139 LUT). GPT-5.1 generates resource-efficient code but tends toward conservative pragma usage, resulting in higher latency on most kernels.

\vskip -14pt\relax
\subsection{Ablation Study}
\vskip -1pt\relax

\begin{figure}[t]
  \centering
  \begin{minipage}{0.48\linewidth}
    \centering
    \includegraphics[width=0.88\linewidth]{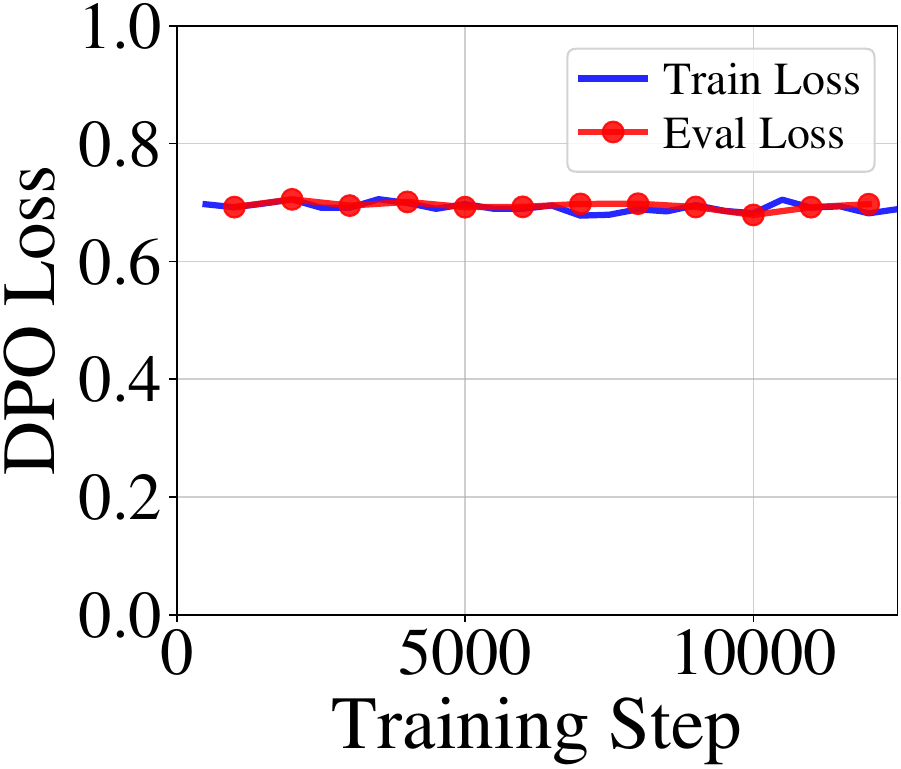}
    \subcaption{DPO training loss}
  \end{minipage}
  \hfill
  \begin{minipage}{0.48\linewidth}
    \centering
    \includegraphics[width=0.88\linewidth]{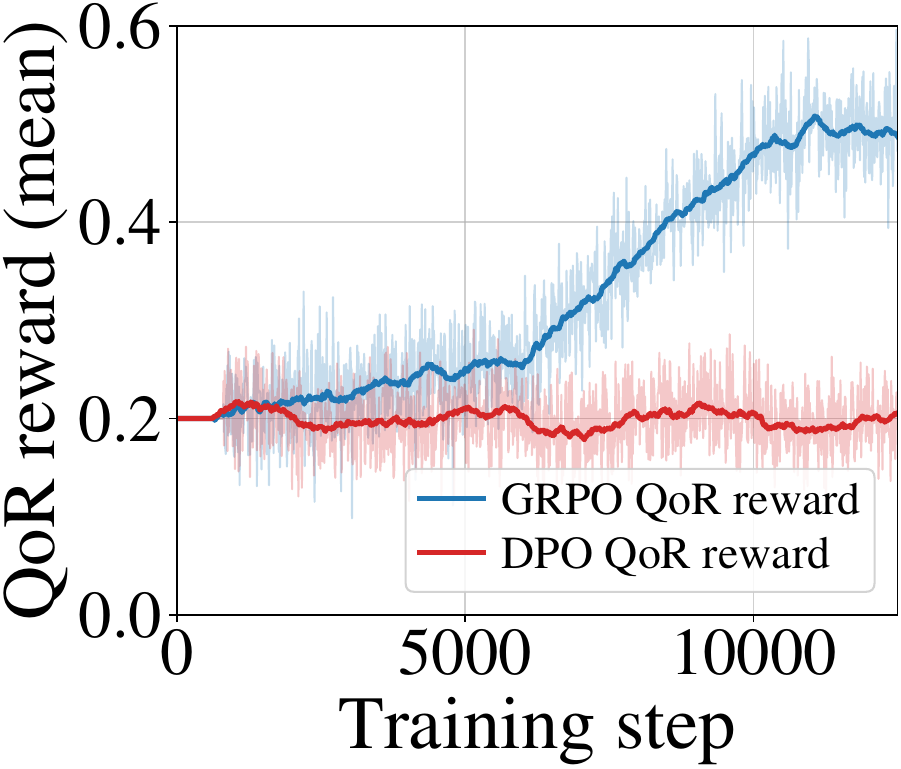}
    \subcaption{QoR reward comparison}
  \end{minipage}
  \Description{(a) DPO loss flat at 0.693. (b) QoR reward: GRPO rises from 0.2 to 0.5 while DPO stays at 0.2.}
  \caption{DPO training loss and QoR reward comparison.}
  \label{fig:dpo_loss}
  \vskip -6pt\relax
\end{figure}

\textbf{Training Strategy.}
PPA-RTL~\cite{zhao2025ppartl} applies Direct Preference Optimization (DPO) for QoR-aware RTL generation. We evaluate whether DPO can similarly improve HLS code quality. Figure~\ref{fig:dpo_loss} contrasts DPO and our GRPO training dynamics. DPO preference pairs are constructed from the same 1,013 applications used for GRPO, each with multiple design variants synthesized with different HLS pragma configurations and annotated with real QoR metrics---latency, BRAM, LUT, DSP, and FF---from Vitis HLS. For each application, we pair the lowest-latency design (chosen) against the highest-latency design (rejected), yielding strong preference signals with large quality gaps. Despite this favorable setup, the DPO loss plateaus around 0.693 (close to $\ln 2$, the random-guess baseline for binary cross-entropy), indicating that DPO fails to learn meaningful preferences between HLS designs. This confirms the limitation: pragma-level differences produce minimal token-level distances that preference learning cannot distinguish. For fair comparison, Figure~\ref{fig:dpo_loss}b evaluates both DPO and GRPO checkpoints using the same QoR reward computation (round-robin pairwise $r_q$ via the proxy reward model). The DPO-trained model's QoR reward remains flat at $\sim$0.2 throughout, confirming that DPO fails to improve hardware quality. In contrast, GRPO's QoR reward rises steadily from 0.2 to $\sim$0.5, demonstrating that the proxy reward provides effective learning signals that drive continuous QoR improvement.

\begin{table}[b]
\centering
\caption{Training cost and QoR comparison: Synthesis-in-the-loop vs.\ HLS-Seek.}
\label{tab:training_cost}
\footnotesize
\setlength{\tabcolsep}{3pt}
\begin{tabular}{lccccc}
\toprule
\textbf{Method} & \textbf{Time/step} & \textbf{Wall-clock} & \textbf{Steps} & \textbf{Low. Lat.} & \textbf{Pareto} \\
\midrule
Real Reward & 180 s & 73h & 1,460 & 3/30 & 2 \\
Proxy (Ours) & 8.8 s + fallback & 73h & 12,500 & 19/30 & 9 \\
\bottomrule
\end{tabular}
\end{table}

\noindent\textbf{Real Reward vs.\ Proxy Reward GRPO.}
To further demonstrate the effectiveness of our proposed proxy reward, we compare it against real synthesis-based reward, which uses synthesis QoR as a reinforcement learning signal, representing a state-of-the-art approach in RTL code generation (e.g., ChipSeek~\cite{chen2025chipseek}). Specifically, we show that this approach is impractical for training high-quality HLS code generation models, as electronic design automation (EDA) synthesis becomes the dominant bottleneck in the training loop. Table~\ref{tab:training_cost} compares GRPO with full Vitis HLS synthesis (Real Reward) against our proxy reward approach on both training cost and final QoR. Under the same 73-hour wall-clock budget, proxy reward completes 12,500 GRPO steps (8.8 s/step for proxy inference plus $\sim$42 hours of selective synthesis fallback triggered by MC dropout uncertainty), while real reward completes only 1,460 steps at 180 s/step. The QoR gap is decisive: Proxy achieves 19/30 lowest-latency kernels and 9 Pareto-dominance wins, versus just 3/30 and 2 for Real Reward. Note that completing all 12,500 steps with real reward would require $\sim$26 days of wall-clock time, making full synthesis-in-the-loop RL impractical.

% The uncertainty-aware switching mechanism (Figure~\ref{fig:ranker_accuracy}b) enables this efficiency: the MC dropout trigger rate starts above 55\% in early training---invoking real synthesis when the proxy is least reliable---and decreases to below 5\% at convergence. Each triggered synthesis result online updates the reward model, creating a self-improving cycle that continuously expands proxy coverage.

\vskip -10pt\relax
\subsection{Case Study}
\vskip -1pt\relax

\begin{figure}[t]
  \centering
  \includegraphics[width=\linewidth,height=100pt]{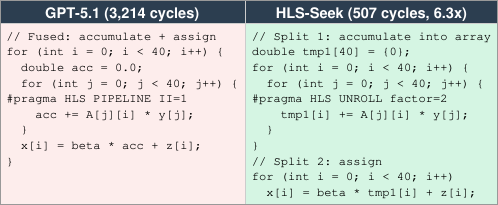}
  \Description{Code comparison between GPT-5.1 and HLS-Seek on GEMVER kernel.}
  \caption{Key code snippets of GEMVER.}
  \label{fig:gemver_case}
  \vskip -6pt\relax
\end{figure}

We illustrate HLS-Seek's advantage through a case study on PolyBench GEMVER (Figure~\ref{fig:gemver_case}), a BLAS matrix-vector computation with $N{=}40$ and \texttt{double} precision. GPT-5.1 generates a fused loop with a scalar accumulator---functionally correct but inherently serial, as the scalar cannot be partitioned for parallel memory access. It applies only \texttt{PIPELINE II=1} on the innermost loop with no array partitioning, yielding low resource usage (22 DSP, 3,202 LUT) but high latency (8,042 cycles). Stage~2, which requires column-wise access to $A$ ($A[j][i]$), suffers memory conflicts that inflate its latency to $\sim$3,214 cycles. HLS-Seek learns to apply \textit{scalar expansion} and \textit{loop fission}: replacing \texttt{double acc} with \texttt{double tmp1[40]} and splitting the loop into independent accumulation and assignment phases. The array form enables \texttt{ARRAY\_PARTITION} for parallel bank access, while loop fission allows independent scheduling of each phase. Combined with \texttt{UNROLL factor=2} on both loop levels, these transformations yield a \textbf{4.1$\times$} overall latency reduction (8,042$\to$1,938 cycles) at the cost of 5.5$\times$ LUT overhead (3,202$\to$17,716). Notably, this is not mere pragma insertion but \textit{algorithmic restructuring}---an optimization pattern that HLS-Seek discovers through reinforcement learning with QoR-aware reward signals. This demonstrates that beyond pragma optimization, HLS-Seek can learn to transform code structure itself to unlock hardware parallelism that is inaccessible to pragma-only approaches.

\vskip -18pt\relax
\section{Conclusion}

We presented HLS-Seek, a QoR-aware framework for NL-to-HLS code generation that combines a three-stage training pipeline with GRPO-based reinforcement learning and a comparative reward model. The uncertainty-aware proxy reward switching mechanism enables online self-improvement of the reward model, achieving 8.5$\times$ training speedup over synthesis-in-the-loop RL while delivering substantially better final QoR under the same wall-clock budget (19/30 vs.\ 3/30 lowest-latency kernels). HLS-Seek achieves 84.7\% syntax correctness pass@1 and 81.4\% functional correctness pass@5 on HLS-Eval with only 7B parameters, surpassing GPT-5.1 on functional pass@5 and leading all baselines at pass@5 and pass@10 on the ForgeHLS test set. On QoR evaluation, HLS-Seek consistently produces lower-latency designs by learning aggressive pragma strategies and algorithmic restructuring that general-purpose LLMs cannot discover. Our results demonstrate that domain-specific RL with learned QoR rewards is a viable path toward fully automated, hardware-efficient HLS code generation.

%%
%% The acknowledgments section is defined using the "acks" environment
%% (and NOT an unnumbered section). This ensures the proper
%% identification of the section in the article metadata, and the
%% consistent spelling of the heading.
\vskip -10pt\relax
\begin{acks}
Supported by the Ministry of Education, Singapore under Tier 3 grant MOE-MOET32024-0003 and Tier 2 grant T2EP20224-0038, and by an AMD Heterogeneous Accelerated Compute Clusters (HACC) hardware donation~\cite{amdHACC}.
\end{acks}

%%
%% The next two lines define the bibliography style to be used, and
%% the bibliography file.
\bibliographystyle{ACM-Reference-Format}
\bibliography{custom}

%%
%% If your work has an appendix, this is the place to put it.

\end{document}